\RequirePackage{fix-cm}

\documentclass{svjour3}                     
\smartqed  
\usepackage{graphicx}
\usepackage{multirow}
\usepackage{booktabs}
\usepackage{float}
\usepackage{amsmath}
\usepackage{authblk}
\usepackage[T1]{fontenc}
\usepackage[normalem]{ulem}
\useunder{\uline}{\ul}{}
\usepackage{longtable}
\usepackage[ruled,vlined]{algorithm2e}
\usepackage{xcolor}

\begin{document}

\title{Efficient Deep Learning Architectures for Fast Identification of Bacterial Strains in Resource-Constrained Devices 
}


\titlerunning{Deep Learning Architectures for Fast Identification of Bacterial Strains}        

\author{Rafael Gallardo Garc\'ia \and
        Sof\'ia Jarqu\'in Rodr\'iguez \and
        Beatriz Beltr\'an Mart\'inez \and
        Carlos Hern\'andez Gracidas \and 
        Rodolfo Mart\'inez Torres 
}

\authorrunning{Rafael Gallardo Garc\'ia et al.} 

\institute{ Rafael Gallardo Garc\'ia \at
                Language and Knowledge Engineering Laboratory, Benem\'erita Universidad Aut\'onoma de Puebla, M\'exico \\
                \email{rafael.gallardo@alumno.buap.mx}            \\
                ORCID: 0000-0001-5085-3501  
            \and
            Sof\'ia Jarqu\'in Rodr\'iguez \at
                Faculty of Chemical Sciences, Benem\'erita Universidad Aut\'onoma de Puebla, M\'exico \\
                \email{ana.jarquin@alumno.buap.mx} \\
                ORCID: 0000-0002-5552-3699
            \and
            Beatriz Beltr\'an Mart\'inez \at
                Language and Knowledge Engineering Laboratory, Benem\'erita Universidad Aut\'onoma de Puebla, M\'exico \\
                \email{bbeltran@cs.buap.mx} \\
                ORCID: 0000-0003-4528-4222
            \and
            Carlos Hern\'andez Gracidas \at 
                Faculty of Physical and Mathematical Sciences, CONACYT-Benem\'erita Universidad Aut\'onoma de Puebla, M\'exico \\
                \email{cahernandezgr@conacyt.mx}
                \\
                ORCID: 0000-0003-0267-6306
            \and
            Rodolfo Mart\'inez Torres \at
                Language and Knowledge Engineering Laboratory, Benem\'erita Universidad Aut\'onoma de Puebla, M\'exico \\
                \email{beetho@cs.buap.mx}
}

\date{Preprint version}

\maketitle

\begin{abstract}
    This work presents twelve fine-tuned deep learning architectures to solve the bacterial classification problem over the Digital Image of Bacterial Species Dataset. The base architectures were mainly published as mobile or efficient solutions to the ImageNet challenge, and all experiments presented in this work consisted in making several modifications to the original designs, in order to make them able to solve the bacterial classification problem by using fine-tuning and transfer learning techniques. This work also proposes a novel data augmentation technique for this dataset, which is based on the idea of artificial zooming, strongly increasing the performance of every tested architecture, even doubling it in some cases. In order to get robust and complete evaluations, all experiments were performed with 10-fold cross-validation and evaluated with five different metrics: top-1 and top-5 accuracy, precision, recall, and F1 score. This paper presents a complete comparison of the twelve different architectures, cross-validated with the original and the augmented version of the dataset, the results are also compared with several literature methods. Overall, eight of the eleven architectures surpassed the 0.95 score in top-1 accuracy with our data augmentation method, being 0.9738 the highest top-1 accuracy. The impact of the data augmentation technique is reported with relative improvement scores.
    
\end{abstract}

\section{Introduction}
    \label{sec:introduction}
    Bacteria are small, unicellular microorganisms which are found almost everywhere on Earth. Most bacteria are not harmful to humans, because less than 1\% of the different species of bacteria make people sick \cite{medline2020}.
	
	Due to the impact of bacteria in human life, the recognition of bacterial genera or species is a very common and important task in many areas such as medicine, veterinary science, biochemistry, food industry, or farming \cite{zielinski2017}. Traditional laboratory methods for the identification of bacterial strains commonly require an expert with knowledge and experience in the field. On the other hand, most techniques designed for rapid and automated identification of microbiological samples are based on biochemical or modular biology technologies \cite{jenkins1991,spratt2004}. These traditional approaches are well established, however they are expensive and time consuming since they require complex sample preparation \cite{ahmed2012}.
	
	Automatizing the process of bacteria identification is very promising in the field of bioimage informatics. This field has yielded powerful solutions for specific image analysis tasks such as object detection, motion analysis or morphometric features \cite{sommer2013}. Even so, most image analysis algorithms have been developed for very specific tasks or biological assays. Faster, more precise and less expensive computational approaches to bacterial strain classification are necessary.
    This paper presents a comparison among several deep learning architectures when solving the task of bacterial strain classification. All experiments were performed with architectures that have been proposed as efficient or mobile solutions to image classification, with the main advantage of low computational cost in terms of memory and number of calculations. In the experiments, the architectures were initialized with a pre-trained model (also called transfer-learning) over the ImageNet dataset \cite{deng2009}, and then fine-tuned to our specific problem. The networks were trained from scratch in the cases where pre-trained models were unavailable. The evaluation of each proposal was performed using 10-fold cross validation over two different datasets: the original one, and a new one, generated by using data augmentation techniques. We present comparative Tables of all architectures, trained and evaluated over both datasets, reporting cross-validation averages of top-1 accuracy, top-5 accuracy, weighted precision, weighted recall, and weighted F1 score.

	The main contributions of this work are summarized as follows:
	\begin{itemize}
	    \item We present proposals that reach and exceed several methods available in the literature when measuring top-1 classification accuracy. \item We carry out a complete and robust evaluation of each method, training and testing with 10-fold cross validation in both datasets, and measuring performance with five different metrics.
	    \item This work indicates that near-state-of-the-art performance can be achieved without using extremely expensive networks: all of our proposals have less than 10 million parameters.
	    \item The efficient and low-resource nature of the networks we used, makes it easy to adapt and scale them to new problems and datasets, also making implementation and deployment on mobile or embedded devices easy.
	    \item Derived from our experiments, this work contains comparative tables of relative improvement, which demonstrate that our data augmentation technique considerably increases the overall performance of several architectures, sometimes with increments of more than 100\%  when comparing against experiments with the original dataset.
	\end{itemize}

\section{State of the art}
    \label{sec:sota}
    Holmberg et al. \cite{holmberg1998bacteria} published the first attempt to classify bacterial strains in an automated way. They used classification trees to extract the most important features and then classified them with artificial neural networks, achieving an accuracy of 76\% in a dataset with 5 species of bacteria. 
	
	Later, in 2001, Liu et al. \cite{liu2001cmeias} presented a computer-aided system to extract size and shape measurements of digital images of microorganisms to classify them into their appropriate morphotype. Their work aimed to classify automatically each cell into one of 11 predominant bacterial morphotypes. This classifier had an accuracy of 96\% on a set of 1,471 cells and 97\% on a set of 4,270 cells. 
	
	Trattner et al. \cite{trattner2004automatic} proposed an automatic tool to identify microbiological data types with computer vision and statistical modeling techniques. Their methodology provided an objective and robust analysis of visual data, to automatize bacteriophage typing methods.
	
	Some works consist in the identification of just one species of bacteria. Forero et al. \cite{forero2006automatic} presented a method for automatic identification of \textit{Mycobacterium tuberculosis} by using Gaussian mixture models. The authors used geometrical and color features of the images. 
	
	In 2013, Ahmed et al. \cite{ahmed2012} published a light scatter-based approach to bacterial classification using distributed computing (due to the computational complexity of their feature extractors). They used Zernike and Chebyshev moments and Haralick texture features, then, the best features were selected using Fisher's discriminant. The classifier was a support vector machine with a linear kernel. The classification accuracy varied from 90\% to 99\% depending on the species.
	
	The Digital Image of Bacterial Species (DIBaS) data\-set was presented in 2017 \cite{zielinski2017}, Zieli\'nski et al. published  a deep learning approach to bacterial colony classification \cite{zielinski2017}. In this work, the authors used Convolutional Neural Networks as feature extractors and support vector machines or random forests as classifiers. Their overall highest accuracy over the DIBaS dataset was 97.24\%, achieved by Fisher vector \cite{perronnin2007fisher} encoders combined with VGG-M \cite{VGGsimonyan2014very} network.
	
	In 2018, Faruk Nasip and Zengin Kenan \cite{nasip2018deep} published another deep learning approach to solve the DI\-B\-aS classification task. The authors used two different architectures, which mainly consist in bacteria classification using a VGG Net-based \cite{VGGsimonyan2014very} approach and an AlexNet-based \cite{AlexNetkrizhevsky2012imagenet} approach, achieving 98.25\% and 97.53\% of classification accuracy, respectively. 
	
	More recently, in 2019, three works over the DIBaS dataset were published. The first one, by Khalifa et al. \cite{khalifa2019deep}, reached a 98.22\% of classification accuracy  with a deep convolutional neural network (CNN) and data augmentation techniques, which they called ``Deep bacteria''. The second one, by Muhammed Talo \cite{talo2019automated}, outperformed ``Deep bacteria'', achieving a 99.2\% of classification accuracy  by fine-tuning a pre-trained model of ResNet-50. The third work \cite{rujichan2019bacteria} reached a 95.09\% of classification accuracy  by fine-tuning a pre-trained version of the MobileNet V2 architecture \cite{mobilenetv2}, and using some data augmentation techniques. 
	
	To the best of our knowledge, just three papers about bacteria classification over the DIBaS dataset were published in 2020. Elaziz et al. \cite{abd2020improved} achieved a 98.68\% of classification accuracy, by using fractional-order orthogonal moments to extract features; also, they proposed a new feature selection method, which they called SSATLBO (Salp Swarm Algorithm + teaching-based learning optimization). In \cite{gallardo2020bacteria}, authors achieved a 94.22\% of classification accuracy, by using a fine-tuned version of a pre-trained MobileNet V2 architecture and data augmentation techniques. The last method, proposed by Satoto et al. \cite{satoto2020auto}, reached a 98.59\% (over a subset of the DIBaS dataset, with just four classes) of classification accuracy by using a custom CNN topology and data augmentation techniques.

\section{Traditional methods for bacterial identification}
    \label{sec:traditional_methods}
    In clinical microbiology laboratories, manual and automated methods are used for bacterial identification. Classification of microorganisms to the species level is one of the most important tasks to microbiologists and other scientists involved in many areas. Phenotypic methods are the standard to identify them. However, they have some limitations for certain types of microorganisms. Molecular methods and proteomics-based methods are complementary to phenotypic methods and allow to overcome some limitations, although their implementation is less usual in some laboratories due to the higher cost and experience required for their application.
	
	\subsection{Phenotypic methods}
	    \label{sec:phenotypic}
    	Traditional bacterial phenotypic identification includes some macroscopic features and biochemical properties such as morphology, hemolysis, culture medium, production of certain metabolites, type of growth in relation to atmosphere, temperature and nutrition. Preliminary identification of most cultivable bacteria is based on such morphological characteristics \cite{franco2019advances}.
    	Manual testing to identify bacteria requires from hours to several days and a large amount of biological material, so that most laboratories use commercial biochemical test panels. The results of these panels are obtained in a short time-period using automated systems \cite{varadi2017methods}.
    		
		The following list contains some of the most common techniques that belong to this category:
		
		\begin{itemize}
		    \item Culture medium.
		    \item Multi-test galleries.
		    \item Automated systems.
		\end{itemize}
		
		\subsubsection{Advantages}
    		These methods are inexpensive and very useful in the clinical laboratory because there are many different culture media for bacterial strains. Currently, biochemical test panels can be read with automated systems and provide results in short time-periods, these methods also allow many microorganisms to be analyzed simultaneously.
		    
		\subsubsection{Disadvantages}
    		Most phenotypic methods  are laborious, require  36–48 hours and are limited to bacteria that are difficult to grow, also these methods consume many reagents and are not always useful to identify the microorganism to the species level.
		
	\subsection{Microscopy techniques}
	    \label{sec:microscopy}
    	Optical microscopy is a simple and fast method that often guides to clinical diagnosis. However, this method is not successful by itself, usually because the identification of microorganisms is limited by factors such as small cells, variation in size or resolution of the optical microscope. Visualization under the microscope is more specific and easier to perform with the use of fluorescent markers. \cite{mishra2016applications}.
	
		The following list contains some of the most common techniques that belong to this category:
		\begin{itemize}
		    \item Transmission Electron Microscopy (TEM).
		    \item Scanning Electron Microscopy (SEM).
		    \item Confocal Laser Scanning Microscopy (CLSM). \end{itemize}
		
		\subsubsection{Advantages}
    		Microscopic observation is useful to reveal morphological features of isolated microorganisms and their diagnosis, mainly in the identification of microorganisms in biofilms. In microscopic techniques, fluorescent dyes are commonly used to get detailed visualization and higher-quality micrographs.
		
		\subsubsection{Disadvantages}
    		Most microscopes are expensive and personnel must be trained to use them correctly, living cells are difficult to differentiate from dead cells in samples and a poorly prepared or contaminated sample reduces contrast and resolution. For these reasons, microscopy alone is not very accurate for microorganism identification.

	\subsection {Proteomics-based methods}
	    \label{sec:proteomics}
    	Proteomics is the study and characterization of the set of proteins expressed by a genome (proteome) \cite{aslam2017proteomics}. Methods based on mass spectrometry (MS) are important tools for microbial typing, they detect the mass-charge ratio $(m/z)$ of the analyte. Particularly, the method known as Matrix Assisted Laser Desorption/Ionization-Time of Flight (MAL\-DI-\-TOF) works by generating a peptide mass fingerprint (PMF) in the form of a spectral profile that is characteristic of each species, the result is compared with existing databases for its identification \cite{singhal2015maldi,sauer2010mass}.
    	
    	\subsubsection{Advantages}
    		MAL\-DI-TOF gets results in a few minutes, the protein spectrum that is characteristic of each microorganism provides an identification at the species level, the cost of the reagents is low and allows to analyze a large number of isolates simultaneously.   
		
		\subsubsection{Disadvantages}
        	The MALDI-TOF equipment is expensive and requires a strong initial invest. The detection is not direct from clinical samples, it also requires frequent calibrations and quality controls, the commercial databases used for the comparison and identification of microorganisms are extensive but still limited.
    	
    \subsection{Molecular methods}
        \label{sec:molecular}
    	Molecular methods are complementary or alternative procedures, which provide high-throughput, more sensitive and discriminatory analysis of microorganisms or genetic polymorphisms through the amplification and detection of specific nucleic acid targets. Some strains do not have a specific feature, the same strain can generate different results in repeated tests, most of the time molecular methods can solve these problems \cite{varadi2017methods,franco2019advances}.
		
		The following list contains some of the most common techniques that belong to this category:
		
		\begin{itemize}
		    \item Real-time PCR.
		    \item Pulse-Field Gel Electrophoresis (PFGE).
		    \item 16S-RNA.
		\end{itemize}
		
		\subsubsection{Advantages}
		    Most of these methods are fast and sensitive, they can identify bacteria that are difficult to grow under laboratory conditions, and are able to detect many pathogens simultaneously.
		 
		\subsubsection{Disadvantages}
		    Molecular methods are expensive and not widely used in routine clinical tests. In some cases, identification at the genus and species level could be confusing or incorrect when the obtained sequence is compared with the sequences of the databases.

\section{Materials and methods}
    \label{sec:materials_methods}
    This section describes the methodologies we used to generate the dataset (the class distribution of the samples is also presented). Details about implementation, fine-tuning, training, and evaluation of  all the deep learning architectures are provided as well.
    
    \subsection{Digital Images of Bacteria Species dataset}
        \label{sec:dibas_dataset}
    	The original version of DIBaS dataset contains a total of 33 species of microorganisms, approximately 20 RGB images (2048$\times$1532 pixels) per each species. It was published as the main dataset of the paper \textit{Deep learning approach to bacterial colony classification} \cite{zielinski2017}. We removed the set of images of \textit{Candida albicans} colonies as it is considered fungi \cite{nobile2015}. The dataset was collected by the Chair of Microbiology of the Jagiellonian University in Krakow. The samples were stained using the Gramm's method. All images were taken with an Olympus CX31 Upright Biological Microscope and an SC30 camera with a 100 times objective under oil-immersion \cite{zielinski2017}. The DIBaS dataset is publicly available \footnote{http://misztal.edu.pl/software/databases/dibas/}.
    
    \subsection{Data augmentation}
    	\label{sec:data_augmentation}
\begin{figure}
            \centering
            \includegraphics[width=0.45\textwidth]{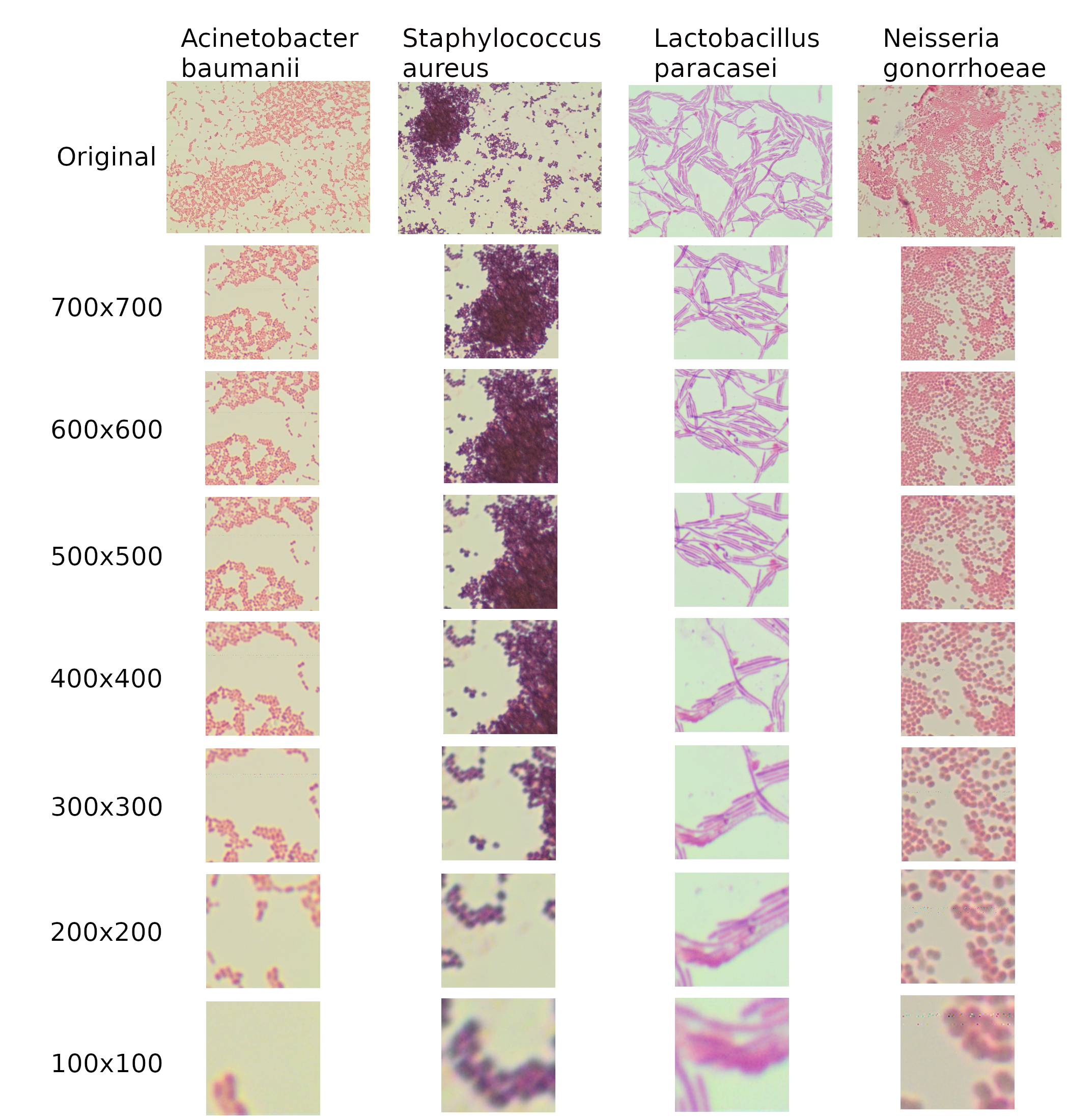}
            \caption{Augmented samples of 4 of the 32 classes in the dataset. This figure shows the original image and one sample of each amplification.}
            \label{fig:examples}
        \end{figure}

\begin{figure}
            \centering
            \includegraphics[width=0.45\textwidth]{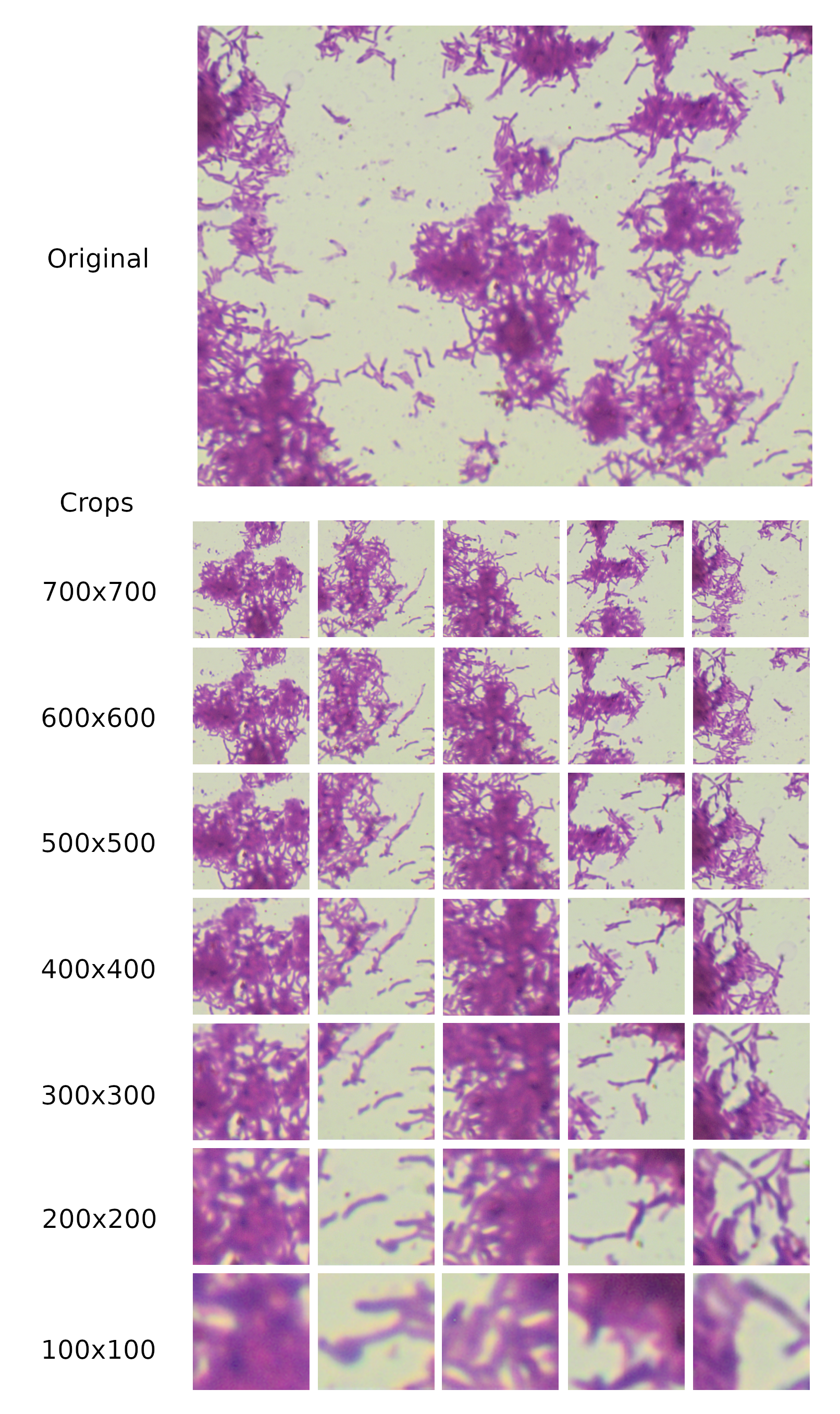}
            \caption{Illustration of each sample generated by our data augmentation method (see Algorithm \ref{algo:augment_data}), applied to one image of the \textit{Actinomyces israelii} class.}
            \label{fig:augmentation_sample}
        \end{figure}

        Our data augmentation strategy is aimed to provide at least 35 different images per each sample in the original dataset. The proposed approach tries to simulate different levels of zoom for the same bacterial sample, which is achieved by cropping different regions of different sizes from the full image. Specifically, we obtained five crops per image, with seven different sizes: from 100 to 700, with steps of 100, as we can see in Algorithm \ref{algo:augment_data}.  All of the proposed architectures are designed to process inputs of 224x224 pixels; hence, after data augmentation, we resized every sample to the input size using Lanczos interpolation. At the end, our augmented version of DIBaS had a total of 24,073 samples (including the original images). Detailed information about sample distribution across classes is available in Table \ref{tab:distribution}. We provide some graphical examples of the samples produced by our data augmentation method: Figure \ref{fig:examples} shows some sample images generated with our data augmentation method and Fig. \ref{fig:augmentation_sample} presents a detailed visualization of the augmentation process for one sample of the \textit{Actinomyces israelii} class.
        
	    \begin{algorithm}[]
            \SetAlgoLined
            \KwResult{An augmented version of the DIBaS dataset.}
            
             \For{Species in DIBaS}
             {
                \For{Sample in Species}
                 {
                     \For{Shape in [100,200,300,400,500,600,700]}
                     {
                        $L$=Obtain 5 crops of Sample with size Shape\;
                        resize all images in L to 224x224 px\;
                     }
                     add all images in L to augmented folder\;
                     resize Sample to 224x224px\;
                     add Sample to augmented folder\;
                 }
             }
             \caption{Data augmentation process.}
             \label{algo:augment_data}
        \end{algorithm}
        
        \begin{table}
            \caption{Samples per species, both in original and augmented versions of the dataset.}
            \label{tab:distribution}       
            \centering
            \resizebox{0.45\textwidth}{!}{%
                \begin{tabular}{@{}rlcc@{}}
                \toprule
                Genera            & Species        & \# Original & \# Augmented \\ \midrule
                Acinetobacter     & baumanii      & 20               & 709               \\
                Actinomyces       & israelii      & 23               & 828               \\
                Bacteroides       & fragilis      & 23               & 828               \\
                Bifidobacterium   & spp.          & 23               & 828               \\
                Clostridium       & perfringens   & 23               & 828               \\
                Enterococcus      & faecalis      & 20               & 720               \\
                                  & faecium       & 20               & 720               \\
                Escherichia       & coli          & 20               & 720               \\
                Fusobacterium     & spp.          & 23               & 828               \\
                Lactobacillus     & casei         & 20               & 720               \\
                                  & crispatus     & 23               & 720               \\
                                  & delbrueckii   & 23               & 720               \\
                                  & gasseri       & 23               & 720               \\
                                  & jehnsenii     & 23               & 720               \\
                                  & johnsonii     & 23               & 720               \\
                                  & paracasei     & 23               & 720               \\
                                  & plantarum     & 23               & 720               \\
                                  & reuteri       & 23               & 720               \\
                                  & rhamnosus     & 23               & 720               \\
                                  & salivarius    & 23               & 720               \\
                Listeria          & monocytogenes & 22               & 792               \\
                Micrococcus       & spp.          & 21               & 756               \\
                Neisseria         & gonorrhoeae   & 23               & 828               \\
                Porphyromonas     & gingivalis    & 23               & 828               \\
                Propionibacterium & acnes         & 23               & 828               \\
                Proteus           & spp.          & 23               & 720               \\
                Pseudomonas       & aeruginosa    & 23               & 720               \\
                Staphylococcus    & aureus        & 23               & 720               \\
                                  & epidermidis   & 23               & 720               \\
                                  & saprophyticus & 23               & 720               \\
                Streptococcus     & agalactiae    & 23               & 720               \\
                Veionella         & spp.          & 22               & 792               \\ \bottomrule
                \end{tabular}%
            }
        \end{table}
        
    \subsection{Deep learning architectures}
        \label{sec:deep_archs}
        This section contains a short description of each architecture we present in this paper. Overall, this paper presents a total of twelve architectures: all are product of transfer-learning and fine-tuning techniques. To accelerate the training process and increase the chances of a good generalization, we used pre-trained models (when available) on ImageNet to initialize the weights of our networks, after fine-tuning we trained the networks end-to-end. Our fine-tuning phase consisted in the modification of some layers of the networks (commonly, the last or the fully-connected layer), by making it suitable to solve a problem with 32 classes (original architectures were designed to solve a 1000-class problem), details about each implementation are described in the corresponding section.
	     
     \subsubsection{EfficientNet}
        \label{sec:efficientnet}
        It was published under the name \textit{EfficientNet: Rethinking Model Scaling for Convolutional Neural Networks} \cite{tan2019efficientnet}. One of the main contributions of the paper was a scaling method that uniformly scales all dimensions of a network (depth, width, and resolution), by using a \textit{compound coefficient}. The compound scaling method uses a compound coefficient $\phi$ to uniform scale the network dimensions. Depth ($d$), width ($w$), and resolution ($r$) of the network can be calculated with $\phi$ as follows:
        
        \begin{equation} \label{eq:efficientnet}
            \begin{split}
                d=\alpha^{\phi}, \\
                w = \beta^{\phi}, \\
                r = \gamma^{\phi} ,\\
            \end{split}
        \end{equation}
        where $\alpha$, $\beta$, and $\gamma$ are constants that can be determined by a small grid search \cite{tan2019efficientnet}. 
        
        The authors proposed a family of models, called \textit{EfficientNets}. The largest model achieved a state-of-the-art score of 84.3\% of top-1 accuracy on ImageNet classification while being 8.4x smaller and 6.1x faster on inference than the best existing ConvNet at that moment \cite{tan2019efficientnet}. Their base architecture is named Efficient\-Net-B0 and contains a total of 5.3 million parameters. Efficient\-Net-B1 to B7 were obtained by using different $\phi$ values on equation \ref{eq:efficientnet}.
        
        We used the implementation and pre-trained models of EfficientNet in PyTorch, provided in the Efficient\-Net-PyTorch repository \footnote{https://github.com/lukemelas/EfficientNet-PyTorch}. The experiments included the three lighter versions of the EfficientNet, i.e., all the versions with less than 10 million parameters. The architectures are named as follows:
        
        \begin{itemize}
            \item efficientnet-b0: EfficientNet-B0, initialized with a pre-trained model on ImageNet.
            \item efficientnet-b1: EfficientNet-B1, initialized with a pre-trained model on ImageNet.
            \item efficientnet-b2: EfficientNet-B2, initialized with a pre-trained model on ImageNet.
        \end{itemize}
        
        All the EfficientNet-based architectures were trained using transfer learning. The original B0, B1, and B2 versions had a total of 5.3, 7.8, and 9.2 million parameters. We performed fine-tuning by modifying the fully connected layer of the network to have 32 output neurons. the final architectures have a total of 4, 6.6, and 7.7 million parameters, respectively.
        
     \subsubsection{MobileNet V2}
        \label{sec:mobilenetv2}
        MobileNet V2 was first reported in the paper \textit{Mobile\-Net\-V2: Inverted Residuals and Linear Bottlenecks} \cite{mobilenetv2}, where the authors proposed a new mobile architecture, based on their main contribution: a novel layer, called inverted residual with linear bottleneck. This novel layer takes a low-dimensional compressed representation as input, and then it is expanded to a high dimension, where it is filtered with a lightweight depth-wise convolution. The base architecture had a total of 3.4 million parameters and achieved a top-1 accuracy score of 72.0\% over the ImageNet classification.
        
        In our experiments, we used the official PyTorch implementation of the MobileNetV2, as well as its pre-trained model on ImageNet. We present just one version of this architecture, called mobilenet\_v2, which is the result of different experiments and combinations of hyperparameters (detailed information available in \cite{gallardo2020bacteria}), from where we selected the best-performing architecture.
        
        We used transfer learning to initialize our architecture; for fine-tuning, we changed the network topology by modifying the last fully connected layer of the original MobileNetV2. In the end, the network had a total of 32 output neurons. After fine-tuning, the total parameter number of the network was reduced to 2.2 millions.

     \subsubsection{MobileNet V3}
        \label{sec:mobilenetv3}
        Andrew Howard et al, presented the next generation of MobileNets (direct sucessor of MobileNetV2) in the paper \textit{Searching for MobileNetV3} in 2019 \cite{howard2019searching}. They tuned MobileNetV3 to be efficient in mobile CPUs with a combination of hardware-aware network architecture search (NAS) and complementary approaches. The authors released two versions: Mobile\-Net\-V3-large and Mo\-bile\-Net\-V3-small, with 5.4 and 2.5 million parameters, respectively, and claimed that, compared to Mobile\-Net\-V2, the largest model is 3.2\% more accurate on ImageNet classification while reducing latency by 15\%. In top-1 classification accuracy of ImageNet, the ``-large'' model achieved 73.8\% and the ``-small'' model achieved a 64.9\%.
        
        Our experiments were performed with a GitHub implementation of MobileNetV3 \footnote{https://github.com/d-li14/mobilenetv3.pytorch}. Code and pre-trained models come from the same source. We used both the large and small version of the network to generate our solutions, which are named as follows:
        
        \begin{itemize}
            \item mobilenet\_v3\_small: MobileNet V3 Small, implemented and trained as described in the original paper, initialized with a pre-trained model on ImageNet.
            \item mobilenet\_v3\_large: MobileNet V3 Large, implemented and trained as described in the original paper, initialized with a pre-trained model on ImageNet.
        \end{itemize}
        
        As with the other architectures, we used transfer-learning to initialize the networks, then, we fine-tuned each version by modifiying the last fully connected layer, in a way that the output is composed by 32 neurons. After fine-tuning, the number of parameters dropped from 5.4 and 2.5 millions, to 4.2 and 1.5  millions, for the large and small models, respectively.
     
     \subsubsection{ShuffleNet V2}
        \label{sec:shufflenetv2}
        First presented in the paper \textit{ShuffleNet V2: Practical Guidelines for Efficient CNN Architecture Design} \cite{ma2018shufflenet}. Their best proposal achieved a top-1 accuracy score of 73.3\% on ImageNet classification.
        
        All the experiments were carried out using the official PyTorch \cite{paszke2017automatic} implementation of this architecture. We used pre-trained models when available on the PyTorch model zoo. The main difference among the four implementations lies in the output channel multiplier, which can be 0.5, 1, 1.5, or 2, which leads the number of parameters of the network to be 1.4, 2.3, 3.5, and 7.4 millions, respectively.
        
        In total, we present experiments with 4 different ShuffleNet V2-based architectures, which are named as follows: 
        \begin{itemize}
            \item shufflenet\_v2\_x0\_5: ShuffleNet V2 with a 0.5 output channel multiplier. This version was initialized with a pre-trained model on ImageNet.
            \item shufflenet\_v2\_x1\_0: ShuffleNet V2 with a 1.0 output channel multiplier. This version was initialized with a pre-trained model on ImageNet.
            \item shufflenet\_v2\_x1\_5: ShuffleNet V2 with a 1.5 output channel multiplier. This version was trained from scratch because no pre-trained models were available.
            \item shufflenet\_v2\_x2\_0: ShuffleNet V2 with a 2.0 output channel multiplier. This version was trained from scratch because no pre-trained models were available.
        \end{itemize}
     
        As we said before, we used transfer-learning in shufflenet\_v2\_x0\_5 and shufflenet\_v2\_x1\_0; the remaining architectures were trained from scratch. For fine-tuning, we changed the network topology by modifying the last fully connected layer of all the architectures in a way that all of them had a total of 32 output neurons (which is the number of classes in our problem). After fine-tuning, the number of parameters of the networks with output channel multipliers of 0.5, 1, 1.5, and 2 was reduced to 0.3, 1.2, 2.5 and 5.5 millions, respectively.
        
     \subsubsection{SqueezeNet}
        \label{sec:squeezenet}
        It appeared for the first time in the paper \textit{SqueezeNet: AlexNet-level Accuracy with 50x Fewer Parameters and $<$0.5 MB Model Size} \cite{iandola2016squeezenet}, as a proposal to provide some of the advantages of small CNNs, such as: less communication across servers during distributed training, less bandwith to export models from the cloud to autonomous cars and to provide CNNs feasible to deploy on field-programmable gate arrays (FPGAs) and other hardware with limited memory. Their base architecture achieved a top-1 accuracy score of 57.5\% on the ImageNet classification task. Their main contribution consisted in a model size reduction up to 510x, when compared to AlexNet \cite{AlexNetkrizhevsky2012imagenet}; also, SqueezeNet models can be saved to disk with a size between 0.47 and 4.8 MB. The vanilla version had a total of 1.248 million parameters. 
        
        Similar to MobileNetV2 and ShuffleNetV2, we used the official PyTorch implementation and the pre-trained models on ImageNet to experiment with the Squeeze\-Net architecture, all available on the PyTorch model zoo. 
        We present two SqueezeNet-based architectures to solve our problem, which are called as follows:
        
        \begin{itemize}
            \item squeezenet1\_0: Vanilla version of Squeeze\-Net, also called SqueezeNet 1.0. This version was initialized with a pre-trained model on ImageNet.
            \item squeezenet1\_1: A slightly modified version of the vanilla architecture, also called SqueezeNet 1.1. This version was initialized with a pre-trained model on ImageNet. SqueezeNet 1.1 has 2.4x less computation and 12,000 less parameters \footnote{https://github.com/forresti/SqueezeNet/}.
        \end{itemize}
        
        After transfer-learning, we fine-tuned both networks by modifying the last convolutional layer, to have a total of 32 output filters therein. After fine-tuning, the total parameters of the network reduced to 0.751 and 0.738 millions in squeezenet1\_0 and squeezenet1\_1, respectively.
    
    \subsection{Experiments}
        \label{sec:experiments}
         Above we have just described our architectures and some details about the implementation, in this section we are providing details about the process of training and evaluation with cross-validation. 
        
        \subsubsection{Loss function, optimizer and hyper-parameters}
            \label{sec:hyperparameters}
            The following list contains relevant information about our experimental setup:
            \begin{itemize}
                \item All architectures were trained by using the cross-entropy loss function (a combination between log softmax and the negative log likelihood loss).
                \item All architectures use the Adam optimizer \cite{kingma2014adam}, with the L2 penalty implemented as proposed in the paper \textit{Decoupled Weight Decay Regularization} \cite{loshchilov2017decoupled}.
                \item All architectures are trained end-to-end, even when we used transfer-learning to initialize the weights. This allows the models to learn more problem-specific details.
                \item Learning rate, weight decay and the number of epochs are adjusted according to the performance of each architecture when training. 
                \item Our cross-validation setup uses 10 folds in every experiment. The dataset is randomized but it uses the same random state in all experiments (to ensure uniform and fair evaluations).
            \end{itemize}
        
        \subsubsection{Software}
            \label{sec:software}
            The following list contains relevant information about the language, libraries and software:
            \begin{itemize}
                \item Dataset scripts, training and evaluation code are fully written in Python 3.
                \item We used PyTorch \cite{paszke2019pytorch} as our main framework.
                \item The training and evaluation process of each architecture is a Jupyter Notebook \footnote{https://jupyter.org/}. Hence, we have two notebooks for each architecture, one for the original dataset and one for the augmented version.
                \item All models are evaluated with five metrics: top-1 and top-5 accuracy, weighted precision, weighted recall and weighted F1 score. For the last three metrics, we used the Scikit-Learn implementation \cite{scikit-learn}.
            \end{itemize}
        
        \subsubsection{Hardware}
            \label{sec:hardware}
            Our main workstation during experiments works with an Intel i7-9750H processor, 20 gigabytes of RAM, and a commercial NVIDIA GTX 1660-Ti with 6 gigabytes of VRAM. All experiments, in training and evaluation phases fitted in just 6 gigabytes of VRAM and less than 10 gigabytes of RAM, ensuring the repeatability of our experiments without needing specific or high-tier hardware.
        
\section{Results}
    \label{sec:results}
    To present a complete comparison, we provide four different tables:
    
    \begin{itemize}
        \item Table \ref{tab:results_orig} contains the results of cross-validation with all our proposals, using the original dataset.
        \item Table \ref{tab:results_augmented} contains the results of cross-validation with all our proposals, using our augmented version of the dataset.
        \item Table \ref{tab:results_comparison} shows a comparison of the number of parameters and Top-1 accuracy among the literature methods and the best of our methods (one for each dataset).
        \item Table \ref{tab:relative_improvement} shows the relative difference (or relative improvement) between metrics obtained before and after using our data augmentation method.
    \end{itemize}
	All values in the three tables are averages of the 10 folds.
	
	\begin{table*}[]
        \centering
        \caption{Cross-validation results of our methods when using the original version of the DIBaS dataset. Models with a $\dagger$ were trained from scratch due to a lack of pre-trained models, the remaining ones were initialized with transfer-learning.}
        \resizebox{\textwidth}{!}{%
        \begin{tabular}{@{}lccccccccc@{}}
        \toprule
        Method                 & \# Parameters & Total samples & \# Folds & \# Epochs per fold & Top-1 Accuracy & Top-5 Accuracy & Precision & Recall & F1 Score \\ \midrule
        shufflenet\_v2\_x0\_5  & 0.374 M       & 669           & 10       & 10                 & 0.8893         & \textbf{0.9954}         & 0.9125    & 0.8893 & 0.8810   \\
        squeezenet1\_1         & 0.738 M       & 669           & 10       & 15                 & 0.5810         & 0.9686         & 0.5675    & 0.5810  & 0.5403  \\
        squeezenet1\_0         & 0.751 M        & 669           & 10       & 15                 & 0.4487         & 0.8895         & 0.4366    & 0.4487 & 0.4047  \\
        shufflenet\_v2\_x1\_0  & 1.286 M       & 669           & 10       & 10                 & 0.9207         & 0.9940         & 0.9412    & 0.9207 & 0.9170   \\
        mobilenet\_v3\_small   & 1.505 M       & 669           & 10       & 10                 & \textbf{0.9341}         & \textbf{0.9954}         & \textbf{0.9548}    & \textbf{0.9341} & \textbf{0.9342}  \\
        mobilenet\_v2          & 2.264 M       & 669           & 10       & 10                 & 0.9237         & 0.9910         & 0.9488    & 0.9237 & 0.9236  \\
        $\dagger$ shufflenet\_v2\_x1\_5 & 2.511 M       & 669           & 10       & 20                 & 0.6307         & 0.9626         & 0.6846    & 0.6307 & 0.6111  \\
        efficientnet-b0        & 4.048 M       & 669           & 10       & 15                 & 0.8938         & 0.9895         & 0.9072    & 0.8938 & 0.8873  \\
        mobilenet\_v3\_large   & 4.243 M       & 669           & 10       & 10                 & 0.9132         & 0.9925         & 0.9311    & 0.9132 & 0.9077  \\
        $\dagger$ shufflenet\_v2\_x2\_0 & 5.543 M       & 669           & 10       & 20                 & 0.6486         & 0.9655         & 0.6827    & 0.6486 & 0.6266  \\
        efficientnet-b1        & 6.654 M       & 669           & 10       & 10                 & 0.8354         & 0.9835         & 0.8579    & 0.8354 & 0.8224  \\
        efficientnet-b2        & 7.746 M       & 669           & 10       & 10                 & 0.8668         & 0.9940         & 0.8837    & 0.8668 & 0.8558  \\ \bottomrule
        \end{tabular}%
        }
        \label{tab:results_orig}
    \end{table*}
    
    As said above, Table \ref{tab:results_orig} contains the results of all experiments performed over the original version of the dataset, which consists of between 20 and 23 samples per class. The results were consistent across the five metrics (i.e. the same model achieved the highest score in all metrics): the mobilenet\_v3\-\_small architecture outperformed the other eleven methods in all metrics, only equaled by shufflenet\_v2\_x0\-\_5 in top-5 accuracy. The second best architecture in top-1 accuracy and F1 score was the mobilenet\_v2, only 0.010 points lower than the best model (in both metrics). The shufflenet\-\_v2\_x1\_0 and mobilenet\_v3\-\_large architectures were close to the first one, only exceeded by 0.013 and 0.02 points in top-1 accuracy, respectively. We had models that showed strong under-fitting (reporting low scores in training and testing) to the problem, such as squeeze\-net1\-\_0, squeeze\-net1\_1, shuffle\-net\-\_v2\_x1\_5, and shuffle\-net\-\_v2\-\_x2\_0, all close or under the 0.6 of top-1 accuracy, surprisingly, all the trained models achieved top-5 accuracies higher than 0.8. Two architectures were trained from scratch, and both reported a clear under-fitting, which we attribute to the lack of data (this set only contains 669 samples in total).
    
    \begin{table*}[]
        \centering
        \caption{Cross-validation results of our methods when using the augmented version of the DIBaS dataset. Models with a $\dagger$ were trained from scratch due to a lack of pre-trained models, the remaining ones were initialized with transfer-learning.}
        \resizebox{\textwidth}{!}{%
        \begin{tabular}{@{}lccccccccc@{}}
        \toprule
        Method                 & \# Parameters & Total samples & \# Folds & \# Epochs per fold & Top-1 Accuracy & Top-5 Accuracy & Precision & Recall & F1 Score \\ \midrule
        shufflenet\_v2\_x0\_5  & 0.374 M       & 24073         & 10       & 10                 & 0.9557         & 0.9976         & 0.9602    & 0.9570  & 0.9571   \\
        squeezenet1\_1         & 0.738 M       & 24073         & 10       & 15                 & 0.9136         & 0.9955         & 0.9210    & 0.9136 & 0.9125   \\
        squeezenet1\_0         & 0.751 M       & 24073         & 10       & 15                 & 0.8882         & 0.9937         & 0.9007    & 0.8882 & 0.8866   \\
        shufflenet\_v2\_x1\_0  & 1.286 M       & 24073         & 10       & 10                 & 0.9635         & 0.9978         & 0.9666    & 0.9635 & 0.9636   \\
        mobilenet\_v3\_small   & 1.505 M       & 24073         & 10       & 10                 & 0.9702         & 0.9975         & 0.9711    & 0.9702 & 0.9703   \\
        mobilenet\_v2          & 2.264 M       & 24073         & 10       & 15                 & 0.9737         & \textbf{0.9980}         & \textbf{0.9745}    & 0.9737 & 0.9737   \\
        $\dagger$ shufflenet\_v2\_x1\_5 & 2.511 M       & 24073         & 10       & 15                 & 0.9001         & 0.9906         & 0.9073    & 0.9001 & 0.9000   \\
        efficientnet-b0        & 4.048 M       & 24073         & 10       & 10                 & 0.9720         & 0.9976         & 0.9728    & 0.9720 & 0.9720   \\
        mobilenet\_v3\_large   & 4.243 M       & 24073         & 10       & 15                 & \textbf{0.9738}         & 0.9977         & \textbf{0.9745}    & \textbf{0.9738} & \textbf{0.9738}   \\
        $\dagger$ shufflenet\_v2\_x2\_0 & 5.543 M       & 24073         & 10       & 15                 & 0.8984         & 0.9972         & 0.9706    & 0.8984 & 0.8977   \\
        efficientnet-b1        & 6.654 M       & 24073         & 10       & 10                 & 0.9661         & 0.9965         & 0.9676    & 0.9661 & 0.9661   \\
        efficientnet-b2        & 7.746 M       & 24073         & 10       & 10                 & 0.9715         & 0.9970          & 0.9723    & 0.9715 & 0.9715   \\ \bottomrule
        \end{tabular}%
        }
        \label{tab:results_augmented}
    \end{table*}

    Table \ref{tab:results_augmented} shows a completely different scenario. In this set of experiments, only two models scored less than 0.9 in top-1 accuracy and all scored higher than 0.9 in top-5 accuracy. These experiments were conducted with a dataset with a total of 24,073 samples, the original data was augmented by using the method described in section \ref{sec:data_augmentation}. In this case, our results remain consistent across all the metrics, the best model (mobilenet\-\_v3\-\_large) achieved the higher accuracy in four of the five metrics: 0.9738 in top-1 accuracy, 0.9745 of precision, 0.0738 in recall, and an F1 score of 0.9738; this model was only outperformed by the mobilenet\_v2 architecture in top-5 accuracy (which also scored 0.9745 in precision). The experiments with the same architectures of Table \ref{tab:results_orig} but with our augmented version of the dataset show a strong increment of the scores, which indicates that our data augmentation techniques are highly effective on this problem. From the twelve models, all achieved a score of 0.9 or higher in precision (nine of the 12 exceeded the 0.95), and all models scored 0.99 or higher in top-5 accuracy, also, eight models surpassed the 0.95 points in top-1 accuracy, recall and F1 score.
    
    \begin{table*}[]
        \centering
        \caption{Comparative table of the number of parameters and top-1 accuracy of the methods available in the literature and the best of our methods for each version of the dataset. A * in the \# Parameters column indicates that the number of parameters was approximated, as the original authors did not report the exact value. Methods with a $\star$ indicate one of our proposals.}
        \resizebox{\textwidth}{!}{%
        \begin{tabular}{@{}llccc@{}}
        \toprule
        Method                                      & Description            & \# Parameters        & Data augmentation & Top-1 Test Accuracy \\ \midrule
        Elaziz et al. \cite{abd2020improved}        & SSATLBO                & Not applicable       & No                & 0.9868              \\
        Khalifa et al. \cite{khalifa2019deep}       & Custom CNN             & 0.541 M *            & Yes               & 0.9822              \\
        $\star$ Our (original)                             & MobileNet v3 Small     & 1.505 M              & No                & 0.9341              \\
        Rujichan et al. \cite{rujichan2019bacteria} & MobileNet v2           & 3.904 M *            & Yes               & 0.9509              \\
        Gallardo et al. \cite{gallardo2020bacteria} & MobileNet v2           & 3.904 M              & Yes               & 0.9422              \\
        $\star$ Our (augmented)                            & MobileNet v3 Large     & 4.243 M              & Yes               & 0.9738              \\
        Muhammed Talo \cite{talo2019automated}      & ResNet50               & 23.573 M *           & No                & 0.9922              \\
        Zieli\'nski et al. \cite{zielinski2017}     & Fisher Vectors + VGG-M & 128.897 – 139.70 M * & No                & 0.9724              \\
        Nasip and Zengin \cite{nasip2018deep}       & VGGNet                 & 128.897 – 139.70 M * & No                & 0.9825              \\ \bottomrule
        \end{tabular}%
        }
        \label{tab:results_comparison}
    \end{table*}
    
    The highest top-1 accuracy score reported in the literature is that presented by Muhammed Talo \cite{talo2019automated}, which scored 0.9922 in the original version of the dataset (using 5 fold cross-validation), his method was mainly based on the fine-tuning of a ResNet50 architecture (with an approximate of 23 million parameters). The second best architecture over the original version of the dataset was that presented by Elaziz et al. \cite{abd2020improved}, which consisted in the SSATLBO method, and scored 0.9868 in top-1 accuracy. Table \ref{tab:results_comparison} contains five literature methods that performed their experiments over the original DIBaS dataset, the SSATLBO method is not a deep learning-based method, thus, we cannot compare its computational cost by measuring the number of parameters, however, from the four remaining methods, our mobilenet\_v3\_small architecture scored 0.9341 in top-1 accuracy with just 1.5 million parameters, while the others achieved 0.9724 \cite{zielinski2017}, 0.9825 \cite{nasip2018deep} and 0.9922 \cite{talo2019automated} scores by using architectures with, approximately, 23, 128, and 128 million parameters.
    
    Table \ref{tab:results_comparison} also presents four literature methods that used an augmented version of the DIBaS dataset to perform their experiments. The highest score belongs to the method presented by Khalifa et al., which achieved a top-1 accuracy score of 0.9822 with a custom CNN of, approximately, 0.5 million parameters, their augmented dataset consisted of 6,600 images in training and 5,940 images in testing. The second highest score belongs to our MobileNet v3 Large model (with 4.2 million parameters), which achieved a 0.9738 of top-1 accuracy, trained and evaluated with 10-fold cross-validation over a dataset with 24,073 samples. The third and fourth scores were both achieved by different versions of the MobileNet v2, with top-1 accuracies of 0.9509 \cite{rujichan2019bacteria} and 0.9422 \cite{gallardo2020bacteria}, those architectures mainly vary in their hyperparameters and the data, the former used an augmented version of the dataset with approximately 1,200 samples, and the last used a dataset with 26,524 samples, both proposals used training and test splits instead of cross-validation.
    
    Table \ref{tab:relative_improvement} contains relative improvements between scores obtained before and after augmenting the DIBaS dataset with Algorithm \ref{algo:augment_data}. The relative improvement of each architecture is calculated as follows:
    \begin{equation}
        r = (\frac{M_a - M_o}{M_o}) \times 100,
    \end{equation}
    where $r$ is the relative improvement, $M_a$ is the score of some architecture in metric $M$ when using the original dataset, and $M_o$ is the score of the same architecture, with the same metric, but when using our data augmentation method.
    
    As we can see in Table \ref{tab:relative_improvement}, all architectures reported a positive relative improvement, which indicates that the data augmentation was effective in all cases. Again, our results are consistent among the five metrics, being squeeze\-net1\_0 the architecture with the highest relative improvements in all metrics, surpassing the 100\% both in Precision and F1 scores, which can be interpreted as doubling the score obtained when only the original data was used. Some other architectures also reported very high relative improvements, such as squeeze\-net1\_1,  shuffle\-net\_v2\_x1\_5, 
    and shuffle\-net\_v2\_x2\_0.
    
    \begin{table*}[]
        \centering
        \caption{The relative improvement of all metrics, for each architecture, before and after data augmentation.}
        \label{tab:relative_improvement}
        \resizebox{\textwidth}{!}{%
        \begin{tabular}{@{}lrrrrr@{}}
        \toprule
        Method                   & \multicolumn{1}{l}{Top-1 Accuracy (\%)} & \multicolumn{1}{l}{Top-5 Accuracy (\%)} & \multicolumn{1}{l}{Precision (\%)} & \multicolumn{1}{l}{Recall (\%)} & \multicolumn{1}{l}{F1 Score (\%)} \\ \midrule
        shufflenet\_v2\_x0\_5    & 7.4                                  & 0.2                                  & 5.2                             & 7.6                          & 8.6                            \\
        squeezenet1\_1           & 57.2                                  & 2.7                                  & 62.2                             & 57.2                          & 68.8                            \\
        squeezenet1\_0           & 97.9                                  & 11.7                                  & 106.2                             & 97.9                          & 119.0                           \\
        shufflenet\_v2\_x1\_0    & 4.6                                  & 0.3                                  & 2.6                             & 4.6                          & 5.0                            \\
        mobilenet\_v3\_small     & 3.8                                  & 0.2                                  & 1.7                             & 3.8                          & 3.8                            \\
        mobilenet\_v2            & 5.4                                  & 0.7                                  & 2.7                             & 5.4                          & 5.4                            \\
        shufflenet\_v2\_x1\_5 & 42.7                                  & 2.9                                  & 32.5                             & 42.7                          & 47.2                            \\
        efficientnet-b0          & 8.7                                  & 0.8                                  & 7.2                             & 8.7                          & 9.5                            \\
        mobilenet\_v3\_large     & 6.4                                  & 0.5                                  & 4.6                             & 6.6                          & 7.2                            \\
        shufflenet\_v2\_x2\_0 & 38.5                                  & 3.2                                  & 42.1                             & 38.5                          & 43.2                            \\
        efficientnet-b1          & 15.6                                  & 1.3                                  & 12.7                             & 15.6                          & 17.4                            \\
        efficientnet-b2          & 12.0                                  & 0.3                                  & 10.0                             & 12.0                          & 13.5                            \\ \bottomrule
        \end{tabular}%
        }
    \end{table*}

\section{Discussion}
    \label{sec:discussion}
    As mentioned before, as well as in other literature works (such as \cite{khalifa2019deep}), the low quantity of samples in the original version of the dataset may cause fitting problems in some architectures. Table \ref{tab:results_orig} shows some examples of systems that under-fitted the problem even when using transfer-learning (the case of squeezenet1\_0 and squeezenet1\_0), other architectures like shufflenet\_v2\-\_x1\_5 and shufflenet\_v2\-\_x2\_0 were unable to fit the problem from scratch and more tests are needed to see if transfer-learning can help to get better results in the original dataset. From our perspective, the main problem lies in the lack of training data as its clear that the gap between scores in Table \ref{tab:results_orig} and Table \ref{tab:results_augmented} for the aforementioned architectures is big. On the other hand, we have those architectures that achieved high scores (0.9 or higher) with the original version of DIBaS (such as mobilenet\_v3\_small, shufflenet\_v2\_x1\_0, mobilenet\-\_v2 and mobilenet\_v3\_large), its easy to see that there is no clear relationship between the network capacity (\# Parameters column) and the top-1 accuracy score. We speculate that this difference in the network generalization is caused by the variations of the information flow inside the topology of each architecture, but further research and analysis is needed. In Table \ref{tab:results_orig}, the three highest top-1 accuracy scores were achieved by networks that consist of between 1 and 2.5 million parameters, and networks with less than 1 million or higher than 2.5 million parameters scored lower values (with the exempt of mobilenet\_v3\_large), this could be a clue that this is a good starting point to train good and computationally cheap models with the original dataset without needing strong pre-processing or hyper-parameter tuning to avoid under-or-over fitting in networks with different capacities. Our special cases of networks trained from scratch bring us some interesting data: both models trained from scratch scored $<$0.7 in top-1 accuracy score, and the remaining ShuffleNet-based architectures achieved higher than 0.8 in the same metric, while having less capacity but making use of pre-trained models to initialize the weights. 
    
    The experiments after the data augmentation process are completely different, and we can describe them as successful. Only two models could not score above 0.9 (squeezenet1\_0 and shufflenet\_v2\_x2\_0) in top-1 accuracy, while all models scored higher than 0.99 in top-5 accuracy. The highest top-1 accuracy was achieved by the mobilenet\_v3\_large model, closely followed by the mobilenet\_v2 model (the former is only 0.0001 points better), the latter was the best model in top-5 accuracy score. As we can see in Table \ref{tab:results_augmented}, the highest scores are from architectures composed of between 1.2 and 7.7 million parameters, which is slightly different to the results of Table \ref{tab:results_orig}. We think that the increment in scores is certainly related to the increment of training data, as more data also means  stronger and bigger architectures have sufficient information to better solve the problem. The aforementioned importance of the data is also present in the two special cases where we trained the networks from scratch (shufflenet\_v2\_x1\_5 and shufflenet\_v2\_x2\_0), even when ShuffleNet-based models were better than MobileNet V2 in top-1 accuracy scores over ImageNet. Here, those models could not surpass the 0.91 of top-1 accuracy (and MobileNet V2 scored 0.9737), also, the other two ShuffleNet-based architectures with less parameters but initializated with pre-trained models achieved $>$0.95 of top-1 accuracy, highlighting the importance and impact of transfer-learning in this problem. 
    
    Table \ref{tab:relative_improvement} clearly shows the importance and impact of the data augmentation method. As mentioned before, all architectures reported an improvement, in every metric. We are going to discuss four architectures that showed very interesting results: squeezenet1\_0, squeezenet1\_1, shufflenet\_v2\_x1\_5, and shufflenet\_\-v2\-\_x2\_0. The first two reported the highest relative improvement in all metrics, while also being two of the lighter architectures (by \# of parameters), both networks have a low number of parameters and thus, they may require more data to generalize well. There is a large difference between the relative improvement reported by these architectures and the remaining ones; we consider that this behavior is due to the low capacity of these two networks. On the other hand, we have the shufflenet\_v2\_x1\_5 and shufflenet\_v2\_x2\_0 architectures, which also reported a high relative improvement, but in this case we have a different situation: both architectures were trained from scratch (i.e. without using transfer learning), so, it is clear that the data augmentation method strongly improved the network generalization and overall performance.
    
    As the task of bacterial classification is not well standardized, it is difficult to present fair comparisons among the literature methods. From the works contained in Table \ref{tab:results_comparison}, four use the original version of the dataset and five use data augmentation, but every work has different training and test sets; also, some papers used training and test splits and others used n-fold cross-validation; these non-standardized practices can be counterproductive when trying to make fair comparisons of the metrics among various methods available in the literature or when trying to declare a new state-of-the-art method. With the aforementioned, we present Table \ref{tab:results_comparison} just as a reference and a general overview of the literature methods compared against our proposals. We believe that well-standardized practices and evaluations for this task (and dataset) will lead researchers to increasingly better and more useful solutions to the bacterial classification problem, in that sense, we performed all experiments with 10-fold cross validation (a practice that we hope will be replicated in future works), which gives reliability and certainty to our results, and also gives us the opportunity to present robust scores (by avoiding the problem of having easy-or-hard subsets). Another good step to standardize the experiments is to standardize the data, therefore, our augmented version of the data can be easily generated by running our data augmentation script over the original data.
    
\section{Conclusions}
    \label{sec:conclusions}
    This work presented a complete comparison of the performance of different efficient and mobile deep learning architectures to solve the problem of classification of bacterial species. The results reported here provide evidence that near-state-of-the-art performance can be achieved with computationally inexpensive methods. This work also provides empirical evidence that our data augmentation technique can be a game-changer when solving this problem. In total, in this work we proposed 12 architectures, a data augmentation strategy and 24 different models that try to solve the problem, all evaluated with five different metrics. Our results with the original version of the dataset go from 0.4487 to 0.9341 in top-1 accuracy, and  from 0.8882 to 0.9738 with the same metric, experimenting with some data augmentation techniques. Overall, this paper presents 5 models that surpassed the 0.95 in top-1 accuracy and 14 that scored above 0.9.
    
    This paper also presented an interesting and strong data-related contribution: our data augmentation meth\-od led all architectures to achieve a positive relative improvement. One architecture doubled its performance on two different metrics, while other three architectures reported relative improvements of more than 30\% in each metric. The data augmentation technique presented here showed evidence that it can help some architectures to achieve performances that are similar to those networks that used transfer learning or to help networks with low capacity to achieve higher scores (or even doubling them).
    
    Since bacterial identification is a sensitive area, which can involve serious diseases, a standardized version of this task is needed, in order to ensure the reliability of the systems that are product of research works like those presented in this paper. Like we mentioned in Section \ref{sec:discussion}, with the purpose of ensuring the consistence and repeatability of the experiments, all the proposals presented in this paper were trained and evaluated using 10-fold cross-validation, which also avoids the problem of selecting easy or hard subsets of the dataset. Our experiments are easy to review and analyze because we are publicly sharing our algorithm to obtain the augmented samples as well as our main code and scripts (see Section \ref{sec:declarations}).
    
    This work provides strong evidence that deep but computationally inexpensive networks can achieve near state-of-the-art scores, and thus, be highly effective to solve this problem in the real world, opening up the possibilities to have embedded implementations or mobile applications for automatic bacterial recognition, as it is in this type of devices where maximum benefit can be obtained from systems such as the ones proposed here.

\section{Declarations}
    \label{sec:declarations}
    \subsection{Funding}
        This work did not receive funding of any kind.
    
    \subsection{Conflicts of interest/Competing interests}
        The authors have declared that no conflict of interests or competing interests exist.
        
    \subsection{Availability of data and material}
        In the Data Availability section of their paper \cite{zielinski2017}, authors of the DIBaS dataset provided the following link to access the data: \\ \textit{http://misztal.edu.pl/software/databases/dibas}
        
    \subsection{Code availability}
        The software requirements, Jupyter notebooks, Python scripts and detailed instructions to repeat our experiments are available on the following GitHub repository: \textit{github.com/gallardorafael/EfficientMobileDL\_Bacterial}
        
    \subsection{Authors' contributions}
        Rafael Gallardo García and Sofía Jarquín Rodríguez contributed to the study conception, design, experimentation and writing of the manuscript. Chemical and biological background of bacterial identification provided in this paper was performed by Sofía Jarquín Rodríguez. Rafael Gallardo García worked on data processing, system design, programming, training and evaluation related tasks. Beatriz Beltrán Martínez and Carlos Hernández Gracidas provided useful and complete feedback to ensure strong and robust research results, also, they contributed with detailed revisions of the paper structure, content and language-quality. Rodolfo Martínez Torres provided feedback and assistance for some engineering-related tasks. All authors read and approved the final manuscript.
        
    \subsection{Ethics approval}
        As the original data was acquired in \cite{zielinski2017}, none of the authors of this work were exposed to dangerous bacteria, biological material or to dangers of any kind. With the aforementioned, we consider that this section is not applicable to this paper.
        
    \subsection{Consent to participate}
        Not applicable.
        
    \subsection{Consent for publication}
        Not applicable.


%
%

\bibliographystyle{spmpsci}      

\end{document}